\pdfoutput=1

\documentclass[11pt]{article}

\usepackage{acl}

\usepackage{times}
\usepackage{latexsym}

\usepackage[T1]{fontenc}

\usepackage[utf8]{inputenc}

\usepackage{microtype}

\usepackage{inconsolata}

\usepackage{amssymb}
\usepackage{bm}
\usepackage{amsmath}
\usepackage{algorithm}
\usepackage{algpseudocode}
\usepackage{booktabs}
\usepackage{multirow}
\usepackage{graphicx}
\usepackage{bbding}
\usepackage{enumitem}
\setlist[itemize]{nosep,leftmargin=*}

\title{READ: Improving \underline{R}elation \underline{E}xtraction from an \underline{AD}versarial Perspective}

\author{\textbf{Dawei Li, William Hogan, Jingbo Shang\thanks{\textsuperscript{\dag} Corresponding author}} \\
  University of California, San Diego \\
  \texttt{\{dal034,whogan,jshang\}@ucsd.edu}
}

\begin{document}
\maketitle
\begin{abstract}
    Recent works in relation extraction (RE) have achieved promising benchmark accuracy; however, our adversarial attack experiments show that these works excessively rely on entities, making their generalization capability questionable. 
To address this issue, we propose an adversarial training method specifically designed for RE. 
Our approach introduces both sequence- and token-level perturbations to the sample and uses a separate perturbation vocabulary to improve the search for entity and context perturbations.
Furthermore, we introduce a probabilistic strategy for leaving clean tokens in the context during adversarial training. 
This strategy enables a larger attack budget for entities and coaxes the model to leverage relational patterns embedded in the context. 
Extensive experiments show that compared to various adversarial training methods, our method significantly improves both the accuracy and robustness of the model. 
Additionally, experiments on different data availability settings highlight the effectiveness of our method in low-resource scenarios.
We also perform in-depth analyses of our proposed method and provide further hints.
We will release our code at https://github.com/David-Li0406/READ.
\end{abstract}

\section{Introduction}

Relation extraction (RE) is an important subtask of information extraction and plays a crucial role in many other natural language processing (NLP) tasks like knowledge base construction~\cite{luan2018multi} and question answering~\cite{sun2021reasoning}.
The goal of RE is to determine the relationship between a head entity and a tail entity.
For example, given the sentence ``\emph{Miettinen hired for WPS champ Sky Blue.}'', the RE models are supposed to predict the relation ``\emph{Employee-Of}'' between the head entity ``\emph{Miettinen}'' and the tail entity ``\emph{Sky Blue}''. 
With the recent advances in pre-trained language model~\cite{kenton2019bert,liu2019roberta} and self-supervised learning~\cite{qin2021erica,hogan2022fine,hogan2022overview} techniques, RE models have achieved promising benchmark accuracy, reaching levels comparable to human performance.



\begin{table}[]\small
\centering
\begin{tabular}{c|l|c}
\hline
    & \multicolumn{1}{c|}{Sentence}                                                      & Prediction                                                       \\ \hline
Org & \begin{tabular}[c]{@{}l@{}}\textcolor[RGB]{76,175,80}{Miettinen} hired for WPS champ\\ \textcolor[RGB]{248,128,32}{Sky Blue}.\end{tabular}  & \begin{tabular}[c]{@{}c@{}}Employee-Of\\ \Checkmark\end{tabular} \\ \hline
Adv & \begin{tabular}[c]{@{}l@{}}\textcolor[RGB]{76,175,80}{Miettinen} hired for WPS champ\\ \textcolor[RGB]{248,128,32}{\underline{Jeez} Blue}.\end{tabular} & \begin{tabular}[c]{@{}c@{}}No-Relation\\ \XSolidBrush \end{tabular}    \\ \hline
\end{tabular}
\caption{An example from SemEval. We use \textcolor[RGB]{76,175,80}{green} color to represent the head entity and \textcolor[RGB]{248,128,32}{orange} color to represent the tail entity. \underline{Underlining} is used for word substitution.}
\label{adv example}
\end{table}

The recent success of RE models sparks a growing interest in conducting more detailed analyses~\cite{han2020more,peng2020learning,zhang2023towards}.
A significant issue that arises in this context is to explore \textit{whether the RE model learns from context or entities for relation prediction}.
Analyzing this problem could reveal the underlying nature of RE models and offer informative insights for their improvement.
To address this issue, various methods are proposed such as information masking~\cite{peng2020learning} and counterfactual analysis~\cite{wang2022should}.
One drawback of these methods is they usually involve removing entities or context in the sample and observing the model's performance with the remaining part.
That enables them to draw the conclusion about how much can the model learn from entity/ context when giving each of them individually.
However, whether the model would prefer to learn from context or entities when both of them are given still remains unclear.
We name this problem \emph{learning preference} in RE.

To address this issue, we propose a novel approach \textbf{READ}, a.k.a. improving \textbf{R}elation \textbf{E}xtraction from an \textbf{AD}versarial perspective.
We begin by introducing the utilization of \textbf{adversarial attacks}~\cite{jin2020bert,garg2020bae} as a means to investigate the model's learning preference and robustness.
Adversarial attacks in NLP are designed to deceive the model by making very few text substitutions.
As the example shown in Table~\ref{adv example}, by replacing the original word ``\emph{Sky}'' with another word ``\emph{Jeez}'', the attack method successfully fools the model into assigning an incorrect label ``\emph{No-Relation}'' to this sample.
Adversarial attacks provide a highly insightful perspective for determining the crucial parts of the sample from the model's viewpoint.
In this particular example, we can conclude that the word ``\emph{Sky}'', as a part of the entity name, is crucial for the model to make accurate predictions.

In our preliminary experiment applying adversarial attacks to RE, we discovered a clear over-dependency on entities within the current RE model.
This is consistent with the previous works~\cite{peng2020learning} that RE models tend to utilize shallow cues from entities to make predictions.
Our analysis revealed that this over-dependency is the underlying cause of the models' vulnerability to adversarial attacks and can also lead to poor generalization in clean samples.
So the key to improving current RE models is to mitigate this over-dependency on entities.
 

One straightforward approach to bolster models' robustness is text substitution.
However, the considerable time cost to generate adversarial samples with the text substitution method constrains it in scaling in large RE datasets~\cite{yoo2021towards}. Also, in our preliminary experiments, we observed a performance drop in the clean test set with text substitution, which has also been reported by previous works~\cite{xu2022towards}\footnote{We put the experiment result and analysis of text substitution in Appendix~\ref{Text Substitution Method}}.
So we shift our focus towards \textbf{virtual adversarial training}~\cite{miyato2016adversarial,madry2018towards}, which applies continuous perturbations at the embedding level during training, rendering it a more refined and efficient approach.
Our method builds upon the advancements of the current adversarial training methods in NLP~\cite{zhu2019freelb,li2021token} and introduces both sequence- and token-level perturbations to the RE sample.
To facilitate perturbation searching, we devise a separate perturbation vocabulary that tracks the accumulated perturbation for entity and context respectively.
Furthermore, we propose a novel probabilistic strategy to encourage the model to leverage relation patterns from the unperturbed context.
Through extensive experiments, we demonstrate the effectiveness of our method on both adversarial and clean test samples. 
We also observe significant improvements in low-resource settings, indicating the great potential of our method in scenarios with limited data.
We conduct a series of in-depth analyses to give more hints about READ.

The contribution of our work could be summarized as follows:
\begin{itemize}
    \item We propose READ, a novel adversarial method to improve current RE models' robustness.
    \item READ adopts adversarial attacks to analyze RE models' learning preferences and expose an obvious over-dependency on entities.
    \item To enhance RE models' generalization, READ utilizes a virtual adversarial training explicit design for RE. Experiments on three mainstream datasets demonstrate the effectiveness of READ.
\end{itemize}


\section{Related Work}

\subsection{Relation Extraction}

Early RE methods employ pattern-based algorithms~\cite{mooney1999relational} or statistical methods~\cite{mintz2009distant,riedel2010modeling,quirk2017distant} to handle relation extraction.
Neural-based RE models~\cite{zhang2015relation,peng2017cross,miwa2016end} emerge with the advancements in deep learning and natural language processing.
Among them, the transformer-based RE models~\cite{shi2019simple} achieve state-of-the-art performance.
To further enhance performance, various self-supervised learning mechanism designs for RE have been proposed~\cite{soares2019matching,qin2021erica,hogan2022fine}.

There are some works that explore applying adversarial training in RE.
\citet{qin2018dsgan} proposes a generative adversarial training framework to address the noisy labeling problem in distantly supervised relation extraction.
\citet{hao2021knowing} adopt adversarial training to address the false negatives problem in relation extraction.
Both \citet{zhang2020relation} and \citet{li2023adversarial} design new adversarial training pipelines to generate augmented samples for RE.
In our work, we propose to analyze and improve RE models from an adversarial perspective to expose and reduce the excessive reliance of the models on entities.

\subsection{Adversarial Attack \& Training}
Text substitution is one of the most commonly used methods in NLP to attack models or generate adversarial samples~\cite{iyyer2018adversarial,ebrahimi2018hotflip}.
It replaces the original word with its synonym based on certain criteria like word embedding similarity~\cite{zang2020word,ren2019generating,jin2020bert} or model infilling~\cite{garg2020bae,li2020bert}.
There are also some works that propose character-level~\cite{gao2018black,li2018textbugger} and phrase-level~\cite{lei2022phrase} substitutions to generate various adversarial samples.
However, those substitution methods are often challenged by the massive space of combinations when searching for the target word to replace, making them time-costly to implement~\cite{yoo2021towards}.

Virtual adversarial training (VAT) methods generate adversarial samples by applying perturbations to the embedding space~\cite{miyato2018virtual}.
This helps VAT become more efficient than traditional text substitution methods.
VAT makes the model more robust under adversarial attacks while also improving the model's performance in clean test samples~\cite{miyato2016adversarial,cheng2019robust}.
To make VAT more effective,~\citet{zhu2019freelb} accumulate perturbation in multiple searching steps to craft adversarial examples.
\citet{li2021token} devise a Token-Aware VAT (TA-VAT) method to allocate more attack budget to the important tokens in the sequence.
While there are some works that apply virtual adversarial training methods to RE for different purposes, we propose an Entity-Aware VAT method explicitly designed for RE to mitigate over-dependency and non-generalization on entities.
We give a more detailed discussion about adversarial attacks and training in NLP in Appendix~\ref{A Detailed Survey}.

\section{Adversarial Attack for RE}
\label{Adversarial Attack for RE}

In this section, we start by analyzing the state-of-the-art (SOTA) RE models' performance under textual adversarial attacks. 
Then, through further analysis, we expose the over-dependency and non-generalization on entities in the current RE models.

\subsection{Attack Settings}
\label{Attack Settings}

We apply adversarial attacks on ERICA~\cite{qin2021erica} and FineCL~\cite{hogan2022fine}, the two SOTA models with RE-specific self-supervised training.
We choose three RE datasets to conduct experiments: SemEval-2010 Task 8~\cite{hendrickx2019semeval}, ReTACRED~\cite{stoica2021re} and Wiki80~\cite{han2019opennre}.
For each dataset, we randomly choose 1,000 test samples to conduct experiments on.
We use different attack methods including BAE~\cite{garg2020bae}, TextFooler~\cite{jin2020bert}, TextBugger~\cite{li2018textbugger} and Projected Gradient Descent (PGD) Attack~\cite{madry2018towards}.
Here, PGD Attack is a white-box attack that utilizes the model's gradient, while the remaining three attacks are black-box attacks.
We use Textattack\footnote{https://github.com/QData/TextAttack} package and follow all the hyper-parameter settings in the original papers.


To evaluate how RE models perform under adversarial attacks, we follow the previous works~\cite{li2021searching,xu2022weight} and report clean accuracy (the model accuracy on clean examples), accuracy under attack (the model accuracy on adversarial examples subjected to a specific attack), and the number of queries (the average number of queries the attacker required to perform successful attacks).
The experiment results are shown in Table~\ref{TextFooler attack}.

To access RE models' learning preferences, we analyze whether tokens in entities would be attacked more than them in context.
If so, that means entities are more important than context in the model's perspective.
For each dataset, We calculate how frequently the adversarial attacks involve the entity (Entity Freq) and the proportion of the perturbed entity in all perturbed tokens (Entity Ratios).
We also report the average proportion of the entity length in the sample for comparison (Entity \%).
The experiment results are shown in Table~\ref{Entity under attack}.

\begin{figure*}[!t]
    \centering
    \includegraphics[width=16cm]{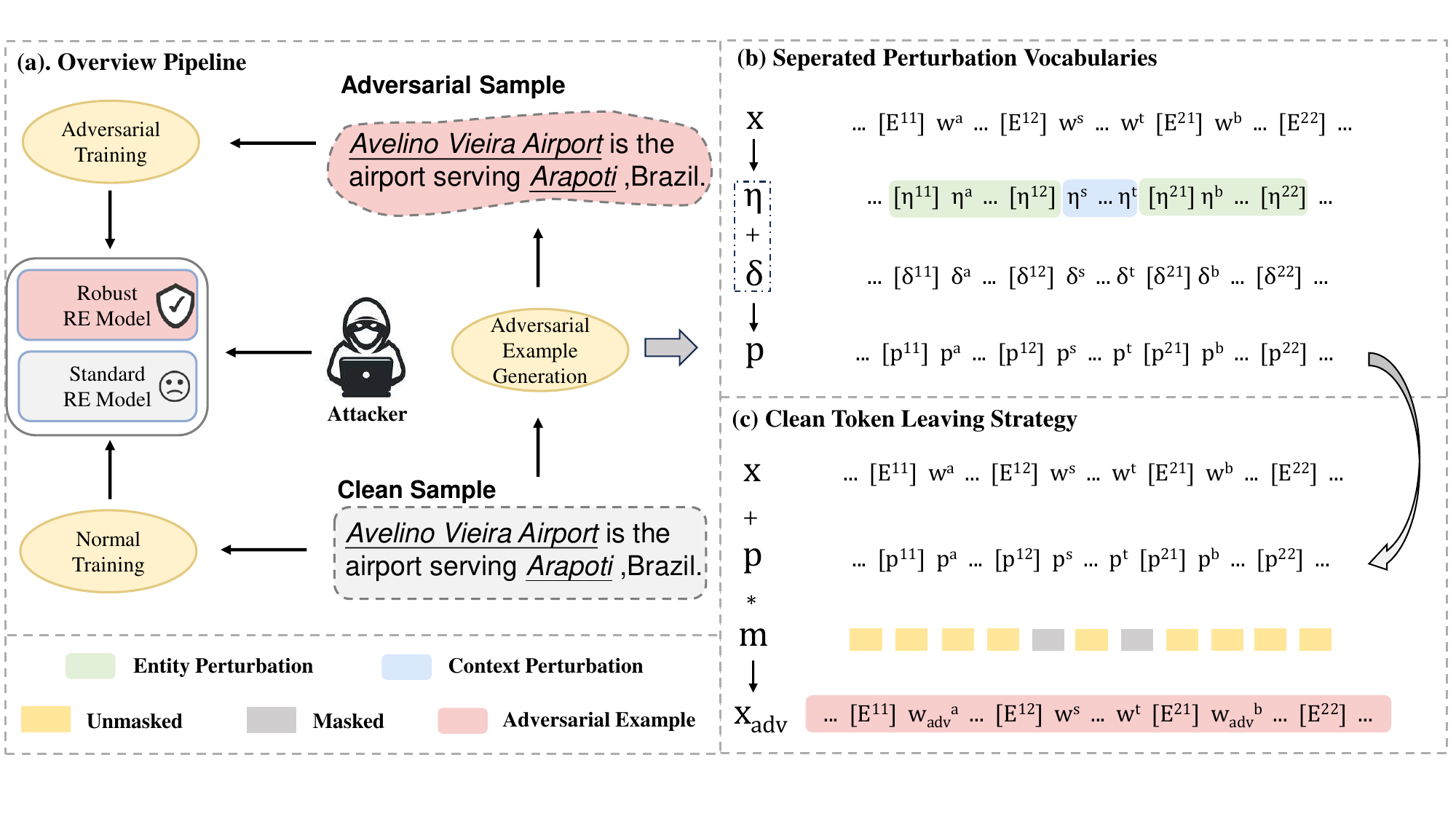}
    \caption{(a) Overview pipeline of our method which adopts adversarial methods to analyze and improve RE models. (b) Separated perturbation vocabularies (Section~\ref{Separate Perturbation Vocabularies}). (c) Clean token leaving strategy (Section~\ref{Probabilistic Clean Token Leaving}). We use ``$[E_{11}]/[E_{12}]$'' and ``$[E_{21}]/[E_{22}]$'' to mark the head and tail entity respectively.}
    \label{fig:pipeline}
\end{figure*}

\subsection{Result Analysis}

Here we analyze the attacking results of TextFooler on FineCL and put the remaining results with other attack methods and models into Section~\ref{Results on Adversarial Samples} and Appendix~\ref{Attack Result on ERICA}.
As shown in Table~\ref{TextFooler attack}, FineCL suffers from a dramatic performance drop up to 91.2\% in the Wiki80 dataset.
In the other two datasets, there is also an obvious performance drop compared with using the clean test set, offering evidence that \textbf{current RE models are not very robust under adversarial attacks}.

\begin{table}[h]\small
\centering
\begin{tabular}{c|ccc}
\hline
Dataset  & Clean & AUA & Query \\ \hline
SemEval  & 92.7  & 18.1 (-80.5\%)  & 73.83   \\ 
ReTACRED & 90.1  & 27.6 (-69.4\%)  & 227.07  \\
Wiki80   & 96.1  & 8.5 (-91.2\%)  & 111.28  \\  \hline
\end{tabular}
\caption{TextFooler attack results on three RE datasets.}
\label{TextFooler attack}
\end{table}

As for the model's learning preference, from Table~\ref{Entity under attack} we find Entity Freq is quite high in the three datasets, suggesting entities are frequently targeted for attacks.
Also, Entity Ratio is much higher than Entity \%, indicating that entities are more often considered important words according to the model's perspective.
Based on these two findings we deduce that \textbf{Current RE models rely more on entities to make predictions.}

The aforementioned conclusion makes us wonder about the RE models' robustness and generalization toward entities.
To evaluate it, we calculate the attack success (AS) rate of entity and context respectively.
As Table~\ref{Entity & context AS} shows, we find the AS of entity is significantly higher than that of context, which means entities are more vulnerable to attacks.
This provides evidence that \textbf{over-dependency on entities has led to a non-generalization within the model.}

\begin{table}[h]\small
\centering
\begin{tabular}{c|ccc}
\hline
         & Entity Freq & Entity Ratio & Entity \% \\ \hline
SemEval  & 77.1        & 38.0         & 12.0     \\ 
ReTACRED & 52.6        & 12.7         & 9.2      \\ 
Wiki80   & 90.7        & 36.4         & 17.4     \\ \hline
\end{tabular}
\caption{Analysis of the model's learning preference. We report how frequently the entity is attacked (Entity Freq), the proportion of the perturbed entity in all perturbed tokens (Entity Ratios), and the average proportion of the entity length in the sample (Entity \%).}
\label{Entity under attack}
\end{table}

\begin{table}[h]\small
\centering
\begin{tabular}{c|cc}
\hline
         & Entity AS & Context AS \\ \hline
SemEval  &  68.5 &  62.3      \\
ReTACRED & 44.2 & 33.9                \\
Wiki80   & 84.2 & 75.5         \\ \hline
\end{tabular}
\caption{Attack success (AS) rate of entity and context. The AS for entity and context is calculated by dividing the total number of successfully attacked entities/contexts by the total number of attacked entities/contexts.}
\label{Entity & context AS}
\end{table}

\section{Adversarial Training for RE}

To improve the robustness and generalization of the RE models, READ employs an \textbf{Entity-Aware Virtual Adversarial Learning} method.
In this section, we first give a brief illustration of the virtual adversarial training (VAT) process, then we will introduce our Entity-Aware VAT method in detail.

\subsection{Virtual Adversarial Training}

In virtual adversarial learning, we first need to find a small perturbation $\bm{\delta}$ that maximizes the misclassification risk of the model.
Then, with the perturbation added to the original inputs $\bm{X}$, the goal of virtual adversarial learning is to optimize the model parameter $\theta$ to minimize the loss of those adversarial samples.
That Min-Max process can be summarized as follows:
\begin{equation}
\label{VAT-Min-Max}
    \min_{\theta} \mathbb{E}_{(\bm{X},y)} \left[ \max_{||\bm{\delta}|| \leq \epsilon} L(f_{\theta}(\bm{X}+\bm{\delta}), y)\right]
\end{equation}
where $\bm{X}$ is the embedding of the input sequence and $y$ is the ground truth label. 
$\epsilon$ is the norm ball used to restrict the magnitude of $\bm{\delta}$.

Commonly, gradient ascent is used to do the perturbation search iteratively since the inner maximize function is non-concave.
At step $t$:
\begin{equation}
\label{VAT-projection}
    \bm{\delta}_{t+1} = \prod_{||\bm{\delta}_{t}||_F < \epsilon} \frac{\bm{\delta}_t + \alpha g(\bm{\delta}_t)}{||g(\bm{\delta}_t)||_F}
\end{equation}
\begin{equation}
\label{VAT-gradient}
    g(\bm{\delta}_t)=\nabla_{\bm{\delta}}L(f_{\theta}(\textbf{X}+\bm{\delta}_t), y)
\end{equation}
where $\prod$ means the process of projecting the perturbation onto the norm ball. In the PGD algorithm, Frobenius norm $F$ is used to constraint $\bm{\delta}$.

\subsection{Separate Perturbation Vocabularies}
\label{Separate Perturbation Vocabularies}

Unlike images in the computer vision field where every pixel only carries limited information across instances, tokens in natural language processing are relatively independent semantic units and different tokens can vary in their importance for the sequence.
Previous work~\cite{li2021token} proposes a Token-Aware VAT method based on this thought and designs a global perturbation vocabulary to record each token's perturbation.

In our work, we borrow this insight and improve it for RE by using separate perturbation vocabularies.
Intuitively, entity and context play quite different roles in the relation extraction process for models~\cite{peng2020learning}.
Entities are the main components for the model to focus on while context can provide auxiliary information.
To address this in adversarial training of RE, we keep two perturbation vocabularies for entities and context separately.

To be specific, we create the entity perturbation vocabulary $\bm{V}_e \in \mathbb{R}^{N \times D}$ and context perturbation vocabulary $\bm{V}_c \in \mathbb{R}^{N \times D}$ at the beginning of the adversarial training.
Here $N$ is the vocabulary size and $D$ is the hidden size of the model's embedding.
In each mini-batch, the $i_{th}$ token in the sequence will be assigned an initialized perturbation from the corresponding vocabulary as the token-level perturbation $\bm{\eta}_{0}^{i}$:

\begin{equation}
\label{vocabulary}
    \bm{\eta}_{0}^{i} = \left\{
\begin{aligned}
\bm{V}_e\left[ w_i\right] & , & w_i \in Entity, \\
\bm{V}_c\left[ w_i\right] & , & w_i \in Context.
\end{aligned}
\right.
\end{equation}
Then we follow \citet{li2021token} exactly to update the token-level perturbation. After the perturbation optimization, the two vocabularies are updated respectively with the token perturbation belonging to their category.

\begin{table*}[]\small
\centering
\begin{tabular}{@{}ccccccccccc@{}}
\hline
\multirow{2}{*}{Dataset}  & \multirow{2}{*}{Method} & \multirow{2}{*}{Clean} & \multicolumn{2}{c}{PGD}                           & \multicolumn{2}{c}{TextBugger}                     & \multicolumn{2}{c}{BEA}                            & \multicolumn{2}{c}{TextFooler}                      \\ \cline{4-11} 
                          &                         &                        & \multicolumn{1}{c}{AUA$\uparrow$}           & Query$\uparrow$         & \multicolumn{1}{c}{AUA$\uparrow$}           & Query$\uparrow$          & \multicolumn{1}{c}{AUA$\uparrow$}           & Query$\uparrow$          & \multicolumn{1}{c}{AUA$\uparrow$}           & Query$\uparrow$           \\ \hline
\multirow{4}{*}{SemEval}  & Normal-Train            & 92.7                   & \multicolumn{1}{c}{42.2}          & 6.55          & \multicolumn{1}{c}{39.2}          & 39.03          & \multicolumn{1}{c}{30.5}           &    75.27            & \multicolumn{1}{c}{15.9}          & 73.83           \\ 
                          & FreeLB                  & \textbf{93.3}          & \multicolumn{1}{c}{45.4}          & 6.80          & \multicolumn{1}{c}{41.5}          & 39.41          & \multicolumn{1}{c}{31.6}          & 75.79          & \multicolumn{1}{c}{15.6}          & 73.35           \\ 
                          & TA-VAT                  & 93.1                   & \multicolumn{1}{c}{45.2}          & 6.75          & \multicolumn{1}{c}{41.6}          & 39.22          & \multicolumn{1}{c}{31.6}          & \textbf{78.93} & \multicolumn{1}{c}{16.5}          & 71.97           \\ 
                          & Ours                    & 93.1                   & \multicolumn{1}{c}{\textbf{51.5}} & \textbf{7.0}  & \multicolumn{1}{c}{\textbf{42.6}} & \textbf{41.18} & \multicolumn{1}{c}{\textbf{32.5}} & 76.7           & \multicolumn{1}{c}{\textbf{18.8}} & \textbf{74.77}  \\ \hline
\multirow{4}{*}{ReTACRED} & Normal-Train            & 90.1                   & \multicolumn{1}{c}{56.4}          & 7.52          & \multicolumn{1}{c}{31.7}          & 89.25          & \multicolumn{1}{c}{41.4}              &     126.27           & \multicolumn{1}{c}{27.6}          & 227.07          \\  
                          & FreeLB                  & 90.0                   & \multicolumn{1}{c}{64.2}          & 7.87          & \multicolumn{1}{c}{29.8}          & 85.83          & \multicolumn{1}{c}{40.1}          & 127.16         & \multicolumn{1}{c}{28.6}          & 228.54          \\ 
                          & TA-VAT                  & \textbf{91.3}          & \multicolumn{1}{c}{68.6}          & 8.11          & \multicolumn{1}{c}{28.9}          & 83.38          & \multicolumn{1}{c}{41.8}          & 128.10         & \multicolumn{1}{c}{30.0}          & 230.88          \\ 
                          & Ours                    & \textbf{91.3}          & \multicolumn{1}{c}{\textbf{76.2}} & \textbf{8.43} & \multicolumn{1}{c}{\textbf{34.0}} & \textbf{89.30} & \multicolumn{1}{c}{\textbf{49.6}}              &    \textbf{140.98}            & \multicolumn{1}{c}{\textbf{38.9}} & \textbf{252.63} \\ \hline
\multirow{4}{*}{Wiki80}   & Normal-Train            & 96.1                   & \multicolumn{1}{c}{58.7}          & 8.34          & \multicolumn{1}{c}{26.3}          & 52.93          & \multicolumn{1}{c}{37.8}              &       46.32         & \multicolumn{1}{c}{8.5}           & 111.28          \\  
                          & FreeLB                  & 95.9                   & \multicolumn{1}{c}{65.3}          & 8.57          & \multicolumn{1}{c}{27.2}          & 53.13          & \multicolumn{1}{c}{39.0}          & 49.1           & \multicolumn{1}{c}{9.0}           & 111.18          \\ 
                          & TA-VAT                  & 96.5                   & \multicolumn{1}{c}{74.0}          & 8.82          & \multicolumn{1}{c}{\textbf{29.2}} & \textbf{54.56} & \multicolumn{1}{c}{39.3}          & \textbf{49.55} & \multicolumn{1}{c}{8.3}           & 107.21          \\ 
                          & Ours                    & \textbf{96.7}          & \multicolumn{1}{c}{\textbf{76.3}} & \textbf{8.99} & \multicolumn{1}{c}{28.8}          & 53.40          & \multicolumn{1}{c}{\textbf{40.0}} & 48.64          & \multicolumn{1}{c}{\textbf{10.7}} & \textbf{112.08} \\ \hline
\end{tabular}
\caption{Experiment results on the three datasets under adversarial attacks. The best results in each dataset are in bold. For each experiment, we run three times and the average scores are reported.}
\label{robust accuracy results}
\end{table*}

\subsection{Probabilistic Clean Token Leaving}
\label{Probabilistic Clean Token Leaving}

To address the importance of entities in adversarial training, we also adopt a probabilistic clean token leaving strategy for context.
In each mini-batch, we randomly choose $n\%$ of tokens $W_c$ in context and mask both their token- and sentence-level perturbation in every perturbation optimization step $t$:
\begin{equation}
\label{strategy1}
    W_c = RandomlySelect(Context, n)
\end{equation}

\begin{equation}
\label{strategy2}
    \bm{X}_{adv}^{i} = \left\{
\begin{aligned}
\bm{X}^{i} & , & w_i \in W_c, \\
\bm{X}^{i} + \bm{\delta}_t + \bm{\eta}_{t}^{i} & , & Otherwise
\end{aligned}
\right.
\end{equation}
There are two benefits of using our probabilistic clean token leaving strategy.
Firstly, the attack budget $\epsilon$ is constant for each sentence, which means reducing context perturbation is equivalent to increasing the attack budget for the entity.
So it serves as an additional attack to further improve the model's robustness and generalization on entities.
This is our main objective given the model's non-generalization and over-dependency on entities.
Also, according to the previous works~\cite{zhang2021understanding,mekala2022lops}, deep neural networks are more willing to learn from clean components with less noise.
So the strategy also gives the model more chances to leverage relational patterns present in the context~\cite{peng2020learning} by learning from those clean tokens.
We give a detailed process of our Entity-Aware VAT method in Figure ~\ref{fig:pipeline}.

\section{Experiment}

In this section, we design experiments to test our Entity-Aware VAT's performance on both clean and adversarial samples.

\subsection{Setup}
To evaluate our method's performance, we report performance on three RE datasets, SemEval-2010 Task 8~\cite{hendrickx2019semeval}, ReTACRED~\cite{stoica2021re} and Wiki80~\cite{han2019opennre}.
We follow the previous work and use 1\%, 10\% and 100\% data in the training set to train the model respectively.
For the baseline RE model, we choose BERT~\cite{kenton2019bert}, RoBERTa~\cite{liu2019roberta}, ERICA~\cite{qin2021erica} and FineCL~\cite{hogan2022fine}.
We choose the two best baseline models, FineCL and ERICA, to apply the adversarial learning methods.
Here we report FineCL's result and put the results of ERICA in Appendix~\ref{Our Method on ERICA}.
We compare our proposed method with FreeLB\cite{zhu2019freelb} and TA-VAT\cite{li2021token}.
They are widely used virtual adversarial learning methods against textual attacks.
For standard accuracy metrics, we follow the previous works and report the F1 score for SemEval and ReTACRED, and the accuracy score for Wiki80.
We also test our method in the document-level RE scenario and put the result in Appeneix~\ref{Our method in Document-level RE}.

We also test our proposed method's performance under adversarial attacks.
All the adversarial attack methods and robustness metrics we use are mentioned in Section~\ref{Attack Settings}

\subsection{Implementation Details}

We build our method based on PyTorch-1.8.1\footnote{https://pytorch.org/} deep learning framework and Transformers-2.5.0\footnote{https://huggingface.co/docs/transformers/index} library.
We follow the hyper-parameter settings in the original paper to reproduce each baseline's result.
To improve the experiments' reliability, we report the average results of the top three adversarial hyper-parameter configurations based on their scores in the development set.
Refer to Appendix~\ref{Training Details} for more detailed settings of our experiments.

\begin{table*}[]\small
\centering
\setlength{\tabcolsep}{0.8mm}{
\begin{tabular}{c|ccccccccc}
\hline
Dataset                                                        & \multicolumn{3}{c}{SemEval} & \multicolumn{3}{c}{ReTACRED} & \multicolumn{3}{c}{Wiki80} \\ \hline
Size                                                           & 1\%     & 10\%    & 100\%   & 1\%     & 10\%    & 100\%    & 1\%     & 10\%   & 100\%   \\ \hline
BERT                                                           & 40.8    & 78.7    & 86.4    & 52.4    & 73.3    & 83.2     & 57.1    & 81.0   & 90.7    \\
Roberta                                                        & 50.0    & 81.6    & 85.8    & 58.2    & 82.5    & 88.7     & 60.7    & 85.4   & 91.3    \\
ERICA                                                          & 50.2    & 82.0    & 88.5    & 64.1    & 83.4    & 87.8     & 71.3    & 86.8   & 91.6    \\
FineCL                                                         & 50.8    & 82.7    & 88.6    & 62.8    & 83.2    & 87.1     & 72.7    & 86.9   & 91.6    \\ \hline
\begin{tabular}[c]{@{}c@{}}FineCL + FreeLB\end{tabular} & 52.0    & 83.2    & 88.8    & 63.1    & 84.0    & 88.4     & 72.6    & 87.1   & 91.8    \\
\begin{tabular}[c]{@{}c@{}}FineCL + TA-VAT\end{tabular} & 52.5    & 83.1    & 89.0    & 64.1    & 84.3    & 88.5     & 73.0    & \bf{87.5}   & 91.8    \\
\begin{tabular}[c]{@{}c@{}}FineCL + Ours\end{tabular}   & \bf{53.2}$_{+4.7\%}$    & \bf{83.3}$_{+0.7\%}$    & \bf{89.2}$_{+0.7\%}$    & \bf{64.4}$_{+2.5\%}$    & \bf{85.0}$_{+2.2\%}$    & \bf{88.7}$_{+1.8\%}$     & \bf{73.3}$_{+0.8\%}$    & 87.3$_{+0.5\%}$   & \bf{92.0}$_{+0.4\%}$  \\
\hline
\end{tabular}}
\caption{Experiment results on clean samples of each dataset. We follow the previous works~\cite{hogan2022fine,qin2021erica} and report the F1 score for SemEval and ReTACRED, and the accuracy score for Wiki80. We also add the quantitative comparison results between our method and the FineCL baseline. For each experiment, we run three times and report the average score.}
\label{stadard accuracy results}
\end{table*}

\subsection{Results on Adversarial Samples}
\label{Results on Adversarial Samples}

We employed FineCL as the baseline and assessed the performance of each adversarial method against different attacks. 
To provide a baseline comparison, we designated the standard model without any adversarial training as "Normal-Train", which is included in the first row of Table~\ref{robust accuracy results}. From the scores reported, we can observe some readily apparent trends: (1). Our method consistently outperforms other adversarial training methods under various attack methods on the three datasets. (2) For the ReTACRED dataset, both FreeLB and TA-VAT exhibit a decrease in performance under the TextBugger attack. In contrast, our method demonstrates robust improvements in both accuracy and query number, showing the resilience of our proposed approach. (3) TextFooler achieves the best attack success rate (AS) result on all three datasets, indicating that current RE models are particularly sensitive to the synonym replacement attack employed by TextFooler.

\subsection{Results on Clean Samples}
\label{Results on Clean Samples}

Table~\ref{stadard accuracy results} presents the results evaluated using the clean samples of each dataset.
It is evident that the utilization of adversarial training methods yields a significant improvement in the performance of the best baseline model (FineCL).
Among the three employed adversarial training methods, our Entity-Aware VAT method stands out by reaching the best score across almost every dataset and availability setting.
That indicates our improved adversarial training method also benefits the RE model in clean test samples.

Moreover, we have observed that adversarial learning exhibits a more pronounced impact in low-resource settings.
For example, the improvement brought by our Entity-Aware VAT method on three datasets with 100\% training data is $0.7\%$, $0.8\%$ and $0.4\%$.
However, it achieves a remarkable $4.7\%$ of performance improvement on SemEval with $1\%$ of training data.
This notable improvement highlights the immense potential of adversarial training methods for RE in scenarios with limited resources.

\section{Further Analysis}

In this section, we conduct further experiments to give in-depth analyses of the mechanism of our proposed method.

\label{Ablation Study}

\begin{table}[h]\small
\centering
\begin{tabular}{c|c|c|ccc}

\hline
                                                           & 1\%           & 10\%          & \multicolumn{3}{c}{100\%}                                             \\ \hline
Metrics                                                    & F1            & F1            & \multicolumn{1}{c}{F1}            & \multicolumn{1}{c}{AUA}  & Query \\ \hline
TA-VAT                                                     & 52.5          & 83.1          & \multicolumn{1}{c}{89.0}          & \multicolumn{1}{c}{16.5} & 71.97 \\ \hline
\begin{tabular}[c]{@{}c@{}}Ours \\  w/o SPV\end{tabular}   & 52.8          & 83.2              & \multicolumn{1}{c}{89.1}             & \multicolumn{1}{c}{\textbf{18.8}}     &   73.63    \\ \hline
\begin{tabular}[c]{@{}c@{}}Ours\\     w/o CTL\end{tabular} & 53.1          & 83.0     & \multicolumn{1}{c}{\textbf{89.2}}              & \multicolumn{1}{c}{16.3}     &   72.51    \\ \hline
Ours                                                       & \textbf{53.2} & \textbf{83.3} & \multicolumn{1}{c}{\textbf{89.2}} & \multicolumn{1}{c}{\textbf{18.8}} & \textbf{74.77} \\ \hline
\end{tabular}
\caption{Ablation study on separate perturbation vocabulary (SPV) and clean token leaving (CTL) strategy using SemEval. The attacker used in 100\% training data availability is TextFooler. We include TA-VAT since it is identical to our method when both SPV and CTL are removed.}
\label{ablation study result}
\end{table}

\subsection{Ablation Study}

The separate perturbation vocabulary (SPV) and clean token leaving (CTL) strategy are the two main methods we propose for adversarial training in RE.
In this section, we conduct an ablation study on them to figure out each method's effectiveness in improving the robustness and accuracy of the model.
We conduct experiments on SemEval with 1\%, 10\% and 100\% training data availability.
We report F1 in all three availability settings and AUA and Query in 100\% training data availability.

Table~\ref{ablation study result} shows the result of our ablation study.
We also report the model's performance with TA-VAT because our method degrades to be TA-VAT without the two methods we propose.
We find both separate perturbation vocabulary and clean token leaving are effective in improving the model's accuracy in clean samples.
And clean token leaving brings a significant improvement in robustness to the model while the model with separate perturbation vocabulary only does not.
That indicates the improvement in robustness of our method is mainly from clean token leaving in the context.

\begin{figure*}[htbp]
    \centering
    \begin{minipage}[t]{0.45\linewidth}
        \centering
        \includegraphics[width=\textwidth]{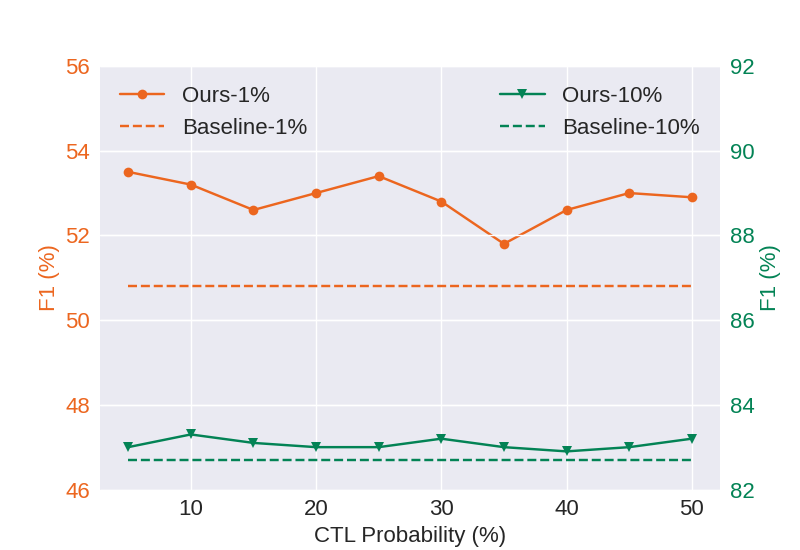}
        \centerline{(a) 1\% and 10\% training data}
    \end{minipage}%
    \begin{minipage}[t]{0.45\linewidth}
        \centering
        \includegraphics[width=\textwidth]{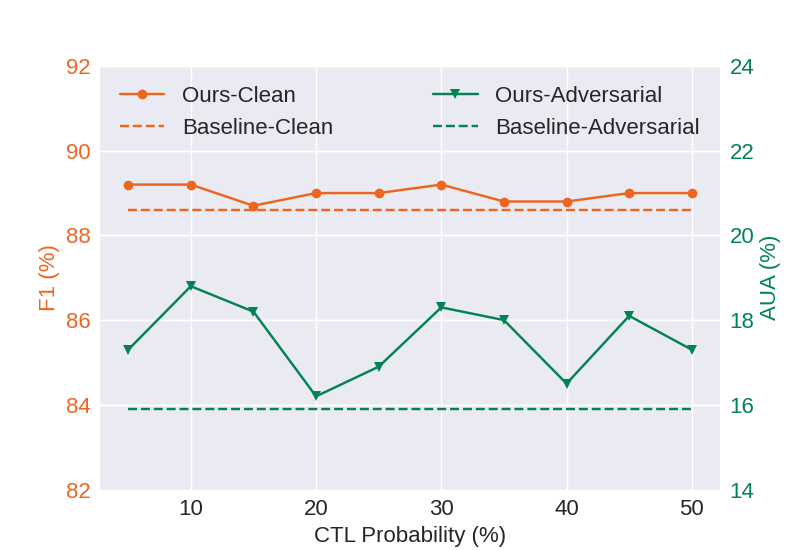}
        \centerline{(b) 100\% training data}
    \end{minipage}
    \caption{Different clean token leaving probability settings in SemEval. For 1\% and 10\% of the training data, we report the F1 score. For the 100\% training data, we report both the F1 score and AUA score}
    \label{fig:CTL}
\end{figure*}

\begin{table}[h!]\tiny
\centering
\begin{tabular}{c|cccc}
\hline
      \begin{tabular}[c]{@{}c@{}}Attack\\ Method\end{tabular}                      &    Method    & Entity Freq & Entity Ratio & Entity AS \\ \hline
\multirow{4}{*}{BAE}        & Normal-Train & 89.0          & 51.1       &  38.2    \\ 
                            & FreeLB       & 91.0          & 53.2       & 36.1    \\ 

                            & TA-VAT       & 89.0          & 51.4       & 36.7    \\ 
                            & Ours         & \bf{87.7}          & \bf{50.4}   & \bf{34.5}        \\ \hline
\multirow{4}{*}{TextFooler} & Normal-Train & 90.7          & 36.4      & 84.2     \\ 
                            & FreeLB       & \bf{89.0}            & 36.7   & 85.2        \\ 
                            & TA-VAT       & 89.7          & 36.5       & 86.9    \\ 
                            & Ours         & 89.7          &\bf{35.4}   & \bf{80.0}        \\ \hline
\end{tabular}
\caption{Adversarial attack results of the entity on Wiki80. BAE and TextFooler are used as attackers.}
\label{entity robustness improvement}
\end{table}

\subsection{Improvement on Robustness of Entity}

Our Entity-Aware VAT method is first introduced to improve the robustness of entities against adversarial attacks.
To investigate its effectiveness in improving entity robustness, we report Entity Freq, Entity Ratio, and Entity AS as we defined in Section~\ref{Adversarial Attack for RE}.
We choose to conduct experiments on the Wiki80 dataset here since it suffers the most from entity attacks, as indicated by the results of our pilot experiments in Section~\ref{Adversarial Attack for RE}.

According to the results presented in Table~\ref{entity robustness improvement}, our method consistently reduces both the frequency of entity attacks and the ratio of perturbed entities compared to the normal-trained baseline and other VAT methods.
This indicates that our method successfully reduces the model's reliance on entities for making predictions.
Also, our method achieves a better performance in terms of entity AS, highlighting its effectiveness in improving the model's robustness toward entities.

\subsection{Impact of Clean Token Leaving Probability}

As we demonstrate in Section~\ref{Ablation Study}, the clean token leaving strategy is a very important design for improving model performance in both clean and adversarial samples.
In this section, we train models with different clean token leaving probabilities to observe their influence on the model performance.
We conduct the analysis on SemEval.

As Figure~\ref{fig:CTL} shows, we add the model without any adversarial training as ``Baseline'' to have a comparison.
It is notable that models with different clean token leaving probabilities consistently outperform baselines.
Additionally, we notice models with different data availability usually achieve the best performance with a relatively small clean token leaving probability (0.05 -- 0.15).

\begin{table}[h!]\small
\centering
\setlength{\tabcolsep}{1.8mm}{
\begin{tabular}{c|ccc}
\hline
Method       & SemEval       & ReTACRED      & Wiki80        \\ \hline
Normal-Train & 51.6          & 62.8          & 72.7          \\ 
w/ DA        & 54.1          & 63.1          & 72.0          \\ 
w/ Ours      & 53.9          & \textbf{64.3} & 73.3          \\ 
w/ DA + Ours & \textbf{55.0} & 64.0          & \textbf{73.5} \\ \hline
\end{tabular}}
\caption{Experiment results with data augmentation on 1\% training data of three datasets. For a fair comparison, we show the result of the optimal model from the development set of our approach.}
\label{da+adv}
\end{table}

\subsection{Comparison and Compatibility with Data Augmentation}

An important finding observed in Section~\ref{Results on Clean Samples} is adversarial training is especially effective in RE when the training data is limited.
Data augmentation~\cite{teru2023semi,hu2023gda} is another widely used technique in low-resource RE.
In this section, we conduct experiments using data augmentation to have a comparison and explore our method's compatibility with data augmentation.
Currently, large language models (LLMs)~\cite{brown2020language,zhang2022opt,anil2023palm,touvron2023llama} with well-designed prompt~\cite{wei2022chain,wang2022self,li2023dail,tong2023eliminating,Tong2024CanLL} show promising performance in generating diverse and high-quality content~\cite{li2024contextualization,tan2024large}.
To benchmark current LLMs' ability in augmenting RE samples, we prompt ChatGPT\footnote{https://platform.openai.com/docs/mode} to do data augmentation.
We put details about the data augmentation method in Appendix~\ref{Data Augmentation with ChatGPT}.

Table~\ref{da+adv} shows the experiment results with 1\% training data.
While data augmentation brings improvement to SemEval and ReTACRED, it also leads to a non-trivial performance drop on Wiki80.
Compared with that, our method consistently improves the model's performance in the three datasets.
Also, combining data augmentation with our method achieves two best results over three datasets, showing our method's compatibility with data augmentation methods.

\section{Conclusion}

In this work, we present READ, a novel method that leverages an adversarial perspective for analyzing and enhancing RE models.
Our adversarial attacks experiment on current SOTA RE models reveals their excessive reliance on entities for relation prediction. 
Through our analysis, this over-dependency is the underlying cause of the models' non-robustness to adversarial attacks and can limit the model's generalization.
To tackle this issue, we propose an Entity-Aware Virtual Adversarial Training method.
Experiment results show our method's effectiveness in improving the performance in both adversarial and clean samples.


\section{Limitations}

This work introduces an Entity-Aware Virtual Adversarial Training method. Similar to other virtual adversarial training algorithms, our method incorporates search perturbation in each mini-batch, leading to a relatively longer training time compared to other normal-trained models. Due to limitations in computing resources, we evaluate our method on four RE datasets, while disregarding scenarios such as continual relation extraction~\cite{han2020continual}, few-shot relation extraction~\cite{gao2019fewrel} and open-world relation extraction~\cite{hogan2023open}. In future research, we plan to investigate the effectiveness of our method in border scenarios.

\bibliography{anthology,custom}

\appendix

\onecolumn

\section{Text Substitution Method}
\label{Text Substitution Method}

In this section, we conduct an experiment using the text substitution method.
Specifically, we follow~\cite{li2020bert} and utilize a BERT model to replace the critical token which can mislead the model most to produce the adversarial samples.
We conduct evaluation using FineCL, on SemEval and ReTACRED with 1\% and 10\% training data.
As Table~\ref{BERT-Attack} shows, while BERT-Attack improves the model's performance on SemEval, it also leads to a non-trivial performance drop on ReTACRED.
This finding aligns with some previous works that point out the traditional text substitution method could cause a performance drop in the clean test set~\cite{yoo2021towards,xu2022towards}.

\begin{table}[h]\small
\centering
\begin{tabular}{l|cccc}
\hline
                                                              & \multicolumn{2}{c}{SemEval}                       & \multicolumn{2}{l}{ReTACRED}                      \\ \cline{2-5}
                                                              & \multicolumn{1}{l}{1\%}           & 10\%          & \multicolumn{1}{l}{1\%}           & 10\%          \\ \hline
FineCL                                                        & \multicolumn{1}{l}{50.8}          & 82.7          & \multicolumn{1}{l}{\textbf{62.8}} & \textbf{83.2} \\ 
\begin{tabular}[c]{@{}l@{}}FineCL\\ +BERT-Attack\end{tabular} & \multicolumn{1}{l}{\textbf{53.1}} & \textbf{83.6} & \multicolumn{1}{l}{62.7}          & 82.7          \\ \hline
\end{tabular}
\caption{Experiment result using BERT-Attack~\cite{li2020bert} on FineCL.}
\label{BERT-Attack}
\end{table}

\section{A Detailed Survey of Adversarial Attack \& Training}
\label{A Detailed Survey}

In the computer vision field, adversarial attacks~\cite{goodfellow2014explaining,carlini2017towards} have been widely explored since it is easy to implement over the continual space of images.
Based on the gradient-based adversarial attacks, various adversarial training~\cite{goodfellow2014explaining,madry2018towards} are proposed.
They add the adversarial sample for the training set to make the model more robust under adversarial attacks.
One major problem of directly applying this gradient-based adversarial training method in NLP is the discrete text prevents the gradient from propagating.

To introduce adversarial training into NLP, some works adopt text substitution as an alternative method to generate adversarial samples~\cite{li2018textbugger,jin2020bert,garg2020bae}.
This method always involves replacing the original word with its synonym based on certain criteria like word embedding similarity~\cite{zang2020word,ren2019generating,jin2020bert} or model infilling~\cite{garg2020bae,li2020bert}.
Another commonly used approach to produce adversarial samples is to generate them with a sequence-to-sequence model~\cite{kang2018adventure,han2020adversarial,la2022king}.

In contrast, virtual adversarial training (VAT) methods generate adversarial samples by applying perturbations to the embedding space~\cite{miyato2018virtual}.
That helps VAT become more efficient than traditional text substitution methods.
VAT makes the model more robust under adversarial attacks while also improving the model's performance in clean test samples~\cite{miyato2016adversarial,cheng2019robust}.
To make VAT more effective,~\citet{zhu2019freelb} accumulate perturbation in multiple searching steps to craft adversarial examples.
\citet{li2021token} devise a Token-Aware VAT (TA-VAT) method to allocate more attack budget to the important tokens in the sequence.
Following them, \citet{xu2022weight} combines weight perturbation with embedding perturbation in training to make the model more robust against text adversarial attacks.
While there are some works that apply virtual adversarial training methods to RE for different purpose~\cite{wu2017adversarial,chen2021empower}, we propose an Entity-Aware VAT method explicitly designed for RE to mitigate over-dependency and non-generalization on entities.

Beyond (virtual) adversarial training, there are also many other techniques proposed as defense mechanisms to adversarial attacks.
For example, some works focus on detecting the adversarial samples and correcting them before inputting them into the language model~\cite{wang2021natural,yang2022robust,li2023text}.
However, our goal in this paper is to improve the RE models' robustness during training.
Such plug-in methods outside the models are not within the scope of our consideration.

\section{Attack Result on ERICA}
\label{Attack Result on ERICA}

We also conduct adversarial attacks on ERICA and put the results in Table~\ref{Attack results on ERICA}.
ERICA exhibited a significant decrease in performance across all attack methods, particularly with TextFooler.
Our analysis of learning preference and entity generalization in ERICA is presented in Table~\ref{Entity under attack on ERICA} and Table~\ref{Entity & context AS on ERICA}. 
The high frequency of successful attacks and their success rate on entities indicates that over-dependency and poor-generalization on entities are ubiquitous in RE models.

\begin{table}[h!]\small
\centering
\begin{tabular}{cccccccccc}
\hline
\multirow{2}{*}{Dataset} & \multirow{2}{*}{Clean} & \multicolumn{2}{c}{PGD}          & \multicolumn{2}{c}{TextBugger}   & \multicolumn{2}{c}{BAE}           & \multicolumn{2}{c}{TextFooler}    \\ \cline{3-10} 
                         &                        & \multicolumn{1}{c}{AUA}  & Query & \multicolumn{1}{c}{AUA}  & Query & \multicolumn{1}{c}{AUA}  & Query  & \multicolumn{1}{c}{AUA}  & Query  \\ \hline
SemEval                  & 93.3                   & \multicolumn{1}{c}{46.1} & 7.44 & \multicolumn{1}{c}{38.1} & 40.47 & \multicolumn{1}{c}{25.3} & 99.74  & \multicolumn{1}{c}{9.1}  & 83.19  \\ \hline
Retacred                 & 89.5                   & \multicolumn{1}{c}{56.8} & 7.87 & \multicolumn{1}{c}{27.0} & 83.90 & \multicolumn{1}{c}{37.2} & 124.21 & \multicolumn{1}{c}{25.2} & 221.05 \\ \hline
Wiki80                   & 96.1                   & \multicolumn{1}{c}{68.0} & 8.56 & \multicolumn{1}{c}{27.7} & 53.95 & \multicolumn{1}{c}{15.7} & 74.61  & \multicolumn{1}{c}{12.1} & 118.96 \\ \hline
\end{tabular}
\caption{Adversarial attack results with ERICA. The attack settings and metrics align with the ones used in~\ref{Attack Settings}.}
\label{Attack results on ERICA}
\end{table}

\begin{table}[h!]\small
\centering
\begin{tabular}{cccc}
\hline
         & Entity Freq & Entity Ratio & Entity \% \\ \hline
SemEval  &   72.7      &    30.8      & 12.0     \\ 
ReTACRED &   55.7      &    13.8      & 9.2      \\ 
Wiki80   &   85.3      &    31.6     & 17.4     \\ \hline
\end{tabular}
\caption{Analysis of ERICA's learning preference with TextFooler.}
\label{Entity under attack on ERICA}
\end{table}

\begin{table}[h!]\small
\centering
\begin{tabular}{ccc}
\hline
         & Entity-AS & Context-AS \\ \hline
SemEval  & 86.0 &  81.8      \\
ReTACRED & 56.0 &  45.5               \\
Wiki80   &  79.5    &   71.6       \\ \hline
\end{tabular}
\caption{Attack success (AS) rate of entity and context on ERICA with TextFooler.}
\label{Entity & context AS on ERICA}
\end{table}

\section{Details of Entity-aware Virtual Adversarial Training}
\label{Details of Entity-aware Virtual Adversarial Training}

We give a detailed algorithm for our Entity-aware Virtual Adversarial Training in Algorithm~\ref{Algo}.

\begin{algorithm*}[h]
\caption{Detailed process of our Entity-Aware Virtual Adversarial Training. We use \textcolor[RGB]{76,175,80}{//} to highlight the important steps.}\label{alg}
\begin{algorithmic}[1]
\Require Training Samples $S={(X=\left[w_0,... w_i, ...\right],y)}$, perturbation bound $\epsilon$, initialize bound $\sigma$, adversarial steps $K$, adversarial step size $\alpha$, model parameter $\theta$, clean token leaving probability $n$
\State $\bm{V}_e \in \mathbb{R}^{N \times D} \leftarrow \frac{1}{\sqrt{D}}U(-\sigma, \sigma)$, $\bm{V}_c \in \mathbb{R}^{N \times D} \leftarrow \frac{1}{\sqrt{D}}U(-\sigma, \sigma)$ \quad \textcolor[RGB]{76,175,80}{// Separate Vocabulary Initialization}
\For{epoch = $1,...,$}
    \For{batch $B \in S$}
        \State $\bm{\eta}_{0}^{i} = \left\{\begin{aligned} \bm{V}_e\left[ w_i\right] & , & w_i \in Entity \\ \bm{V}_c\left[ w_i\right] & , & w_i \in Context \end{aligned} \right.$ \quad \textcolor[RGB]{76,175,80}{// Separate Token-level Perturbation Initialization}
        \State $\bm{\delta}_0 \leftarrow \frac{1}{\sqrt{D}}U(-\sigma, \sigma)$, $\bm{g}_0 \leftarrow 0$
        \State $W_c = RandomlySelect(Context, n)$ \quad  \textcolor[RGB]{76,175,80}{// Clean Token Leaving in Context}
        \For{t = $1,...,K$}
            \State $\bm{X}_{adv}^{i} = \left\{\begin{aligned}\bm{X}^{i} & , & w_i \in W_c, \\ \bm{X}^{i} + \bm{\delta}_t + \bm{\eta}_{t}^{i} & , & Otherwise\end{aligned}\right.$
            \State $\bm{g}_t \leftarrow \bm{g}_{t-1} + \frac{1}{K} \mathbb{E}_{(X,y) \in B} \left[ \nabla_{\theta} L(f_{\theta}(\bm{X}_{adv}), y)\right]$
            \State $\bm{g}_{\eta}^{i} \leftarrow \nabla_{\eta^{i}} L(f_{\theta}(\bm{X}_{adv}), y)$
            \State $\bm{\eta}_{t}^{i} \leftarrow n_i * (\bm{\eta}_{t-1}^{i}+\alpha \cdot \bm{g}_{\eta}^{i})/||\bm{g}_{\eta}^{i}||_F)$
            \State $\bm{\eta}_{t} \leftarrow \prod_{||\bm{\eta}||_F < \epsilon}(\bm{\eta}_t)$
            \State $\bm{g}_{\delta} \leftarrow \nabla_{\delta} L(f_{\theta}(\bm{X}_{adv}), y)$
            \State $\bm{\delta}_{t} \leftarrow \prod_{||\bm{\delta}||_F < \epsilon}(\bm{\delta}_{t-1}+\alpha \cdot \bm{g}_{\delta})/||\bm{g}_{\delta}||_F)$
        \EndFor
        \State $\bm{V}_e\left[ w_i\right] \leftarrow \eta_K^i$, $w_i \in Entity$ \quad \textcolor[RGB]{76,175,80}{// Entity Vocabulary Update}
        \State $\bm{V}_c\left[ w_i\right] \leftarrow \eta_K^i$, $w_i \in Context$ \quad \textcolor[RGB]{76,175,80}{// Context Vocabulary Update}
        \State $\theta \leftarrow \theta - g_K$
    \EndFor
\EndFor
\end{algorithmic}
\label{Algo}
\end{algorithm*}

\section{Our Method on ERICA}
\label{Our Method on ERICA}

The performance of our method with ERICA is presented in Table~\ref{clean sample erica}. 
It is evident that with our method, ERICA also demonstrates a non-trivial improvement in each data availability across three RE datasets.

\begin{table}[h!]\small
\centering
\begin{tabular}{c|ccccccccc}
\hline
Dataset                                                        & \multicolumn{3}{c}{SemEval} & \multicolumn{3}{c}{ReTACRED} & \multicolumn{3}{c}{Wiki80} \\ \hline
Size                                                           & 1\%     & 10\%    & 100\%   & 1\%     & 10\%    & 100\%    & 1\%     & 10\%   & 100\%   \\ \hline
ERICA                                                 & \multicolumn{1}{c}{50.2}          & \multicolumn{1}{c}{82.0}          & 88.5          & \multicolumn{1}{c}{64.1}          & \multicolumn{1}{c}{83.4}          & 87.8          & \multicolumn{1}{c}{71.3}          & \multicolumn{1}{c}{86.8}          & 91.6          \\ \hline
\begin{tabular}[c]{@{}c@{}}ERICA\\ +Ours\end{tabular} & \multicolumn{1}{c}{\textbf{51.8}} & \multicolumn{1}{c}{\textbf{82.6}} & \textbf{89.1} & \multicolumn{1}{c}{\textbf{64.6}} & \multicolumn{1}{c}{\textbf{84.8}} & \textbf{88.8} & \multicolumn{1}{c}{\textbf{71.6}} & \multicolumn{1}{c}{\textbf{87.0}} & \textbf{91.8} \\ \hline
\end{tabular}
\caption{Experiment results of ERICA on clean samples of each dataset.}
\label{clean sample erica}
\end{table}

\section{Our method in Document-level RE}
\label{Our method in Document-level RE}

To demonstrate the compatibility of our proposed entity-aware VAT method across various RE scenarios, we conduct an experiment in a document-level RE dataset, Re-DocRED~\cite{tan2022revisiting} and report the results in Table~\ref{tab: doc}.

\begin{table*}[h]
\centering
\begin{tabular}{lcc}
\hline
          & Ign-F1         & F1             \\
          \hline
ATLOP*     & 76.94          & 77.73          \\
DocuNet*   & 77.27          & 77.92          \\
KD-DocRE*  & 77.63          & 78.35          \\
DREEAM*    & 79.66          & 80.73          \\
PEMSCL*    & 79.01          & 79.86          \\
\hline
AA        & 80.39          & 81.34          \\
AA + Ours & \textbf{81.21} & \textbf{82.22} \\
\hline
\end{tabular}
\caption{Experimental results on Re-DocRED dataset. We apply our entity-aware VAT method on AA~\cite{lu2023anaphor} and * denote the results we take from~\citet{lu2023anaphor}.}
\label{tab: doc}
\end{table*}

\section{Training Details}
\label{Training Details}


In our method, we have set the clean token leaving probability to 10\% for SemEval and 15\% for ReTACRED and Wiki80 datasets. Following the approach of ~\citet{hogan2022fine}, the compared models employ the following settings: a batch size of 64, a maximum sequence length of 100, a learning rate of 5e-5, an Adam epsilon of 1e-8, a weight decay of 1e-5, a maximum gradient norm of 1.0, 500 warm-up steps, and a hidden size of 768.
To account for different data availability scenarios, we utilize dropout rates of 0.2/0.1/0.35 and set the maximum number of training epochs to 80/20/8 for training proportions of 0.01/0.1/1.0, respectively.

For all the adversarial training methods, we search adversarial learning rate in [2e-2, 5e-2, 1e-1], attack budget in [2e-1, 4e-1, 6e-1], and perturbation searching steps in [1,2,3].
For each experiment, we employ grid search\footnote{https://wandb.ai/} to discover the above hyperparameters, and we report the average results of the top three configurations based on their scores in the development set.

We train all models on a single A6000 GPU with CUDA version 11.1. 
The training time for a RE model ranges from approximately 20 to 60 minutes, depending on the specific dataset and availability settings.

\section{Data Augmentation with ChatGPT}
\label{Data Augmentation with ChatGPT}

We use the model `GPT-3.5-turbo-0301' to generate augmented data for 1\% training data availability of each dataset.
For each sample, we randomly choose other two samples with the same relation labels and input them into the model as demonstrations.
After getting output from ChatGPT, we verify that the sentence includes both entities mentioned.
If not, we discard the generated output.
We provide an example of the prompt we use in Table~\ref{chatgpt prompt}.

\begin{table}[t]\small
\centering
\begin{tabular}{c|l}
\hline
Prompt & \begin{tabular}[c]{@{}l@{}}Read the following examples of the relation 'Component-Whole(e2,e1)' between the head and tail and write \\ another new example following the same format. Note that the sentence must contain both head and tail:     \\ head: kangaroo, tail: legs, sentence: the kangaroo moves by hopping on its hind legs using its tail for steering \\ and balancing while hopping at speed up to 40mph/60kmh.\\ head: cottage, tail: kitchen, sentence: the cottage kitchen is on the first floor and is fully fitted with fridge,\\ dishwasher, microwave and all the standard self catering facilities.\\ head: armature, tail: coil, sentence: the armature has a coil of wire wrapped around an iron core.\end{tabular} \\ \hline
\end{tabular}
\caption{An example of the prompt we use to generate augmented samples.}
\label{chatgpt prompt}
\end{table}

\end{document}